\documentclass[conference,letterpaper,10pt]{ieeeconf}
\IEEEoverridecommandlockouts

\usepackage{epsfig}
\usepackage{amsmath} %
\usepackage{amssymb}  %
\usepackage{cite}
\usepackage{epstopdf}
\usepackage{caption}
\usepackage{subcaption}
\usepackage{lscape}
\usepackage{hyperref}
\usepackage[dvipsnames]{xcolor}
\usepackage{color}
\usepackage{parskip}
\usepackage{url}
\usepackage{array}
\usepackage{slashbox}
\usepackage{cleveref}
\newcolumntype{C}[1]{>{\centering\arraybackslash}p{#1}}

\DeclareMathOperator*{\argmin}{arg\,min}

\newtheorem{definition}{Definition}

\definecolor{teal}{RGB}{0, 161, 115}
\definecolor{brightBlue}{RGB}{0, 114, 207}
\definecolor{wine}{RGB}{207, 0, 48}
\definecolor{darkgrey}{RGB}{60, 60, 60}

\newcommand{\algname}{\mbox{interaction-dynamics-aware perception zones}}

\newcommand{\xA}{x_\mathrm{E}}         %
\newcommand{\xB}{x_\mathrm{C}}         %
\newcommand{\yA}{y_\mathrm{E}}         %
\newcommand{\yB}{y_\mathrm{C}}         %
\newcommand{\vA}{v_\mathrm{E}}         %
\newcommand{\vB}{v_\mathrm{C}}         %
\newcommand{\psiA}{\psi_\mathrm{E}}         %
\newcommand{\psiB}{\psi_\mathrm{C}}         %
\newcommand{\aA}{a_\mathrm{E}}         %
\newcommand{\aB}{a_\mathrm{C}}         %
\newcommand{\deltaA}{\delta_\mathrm{E}}         %
\newcommand{\deltaB}{\delta_\mathrm{C}}         %
\newcommand{\LA}{d_\mathrm{E}}         %
\newcommand{\LB}{d_\mathrm{C}}         %

\newcommand{\xrel}{x_\mathrm{R}}         %
\newcommand{\yrel}{y_\mathrm{R}}         %
\newcommand{\psirel}{\psi_\mathrm{R}}         %

\newcommand{\uA}{u_\mathrm{E}}    %
\newcommand{\uB}{u_\mathrm{C}}    %

\newcommand{\UA}{\mathcal{U}^\mathrm{E}}    %
\newcommand{\UB}{\mathcal{U}^\mathrm{C}}    %

\begin{document}

\title{Interaction-Dynamics-Aware Perception Zones for\\Obstacle Detection Safety Evaluation}

\author{Sever Topan, Karen Leung, Yuxiao Chen, Pritish Tupekar,\\Edward Schmerling, Jonas Nilsson, Michael Cox, Marco Pavone%
\thanks{The authors are with NVIDIA, {\tt\small \{stopan, kaleung, yuxiaoc, ptupekar, eschmerling, jonasn, mbc, mpavone\}@nvidia.com}}}%

\newcommand{\klnote}[1]{\textcolor{blue} {KL:#1}}
\newcommand{\stnote}[1]{\textcolor{red} {ST:#1}}

\maketitle

\begin{abstract}
To enable safe autonomous vehicle (AV) operations, it is critical that an AV's obstacle detection module can reliably detect obstacles that pose a safety threat (i.e., are safety-critical).
It is therefore desirable that the evaluation metric for the perception system captures the safety-criticality of objects. Unfortunately, existing perception evaluation metrics tend to make strong assumptions about the objects and ignore the dynamic interactions between agents, and thus do not accurately capture the safety risks in reality.
To address these shortcomings, we introduce an \textit{interaction-dynamics-aware} obstacle detection evaluation metric by accounting for \textit{closed-loop dynamic interactions} between an ego vehicle and obstacles in the scene.
By borrowing existing theory from optimal control theory, namely Hamilton-Jacobi reachability, we present a computationally tractable method for constructing a ``safety zone'': a region in state space that defines where safety-critical obstacles lie for the purpose of defining safety metrics.
Our proposed safety zone is mathematically complete, and can be easily computed to reflect a variety of safety requirements.
Using an off-the-shelf detection algorithm from the nuScenes detection challenge leaderboard, we demonstrate that our approach is computationally lightweight, and can better capture safety-critical perception errors than a baseline approach.

\end{abstract}

\section{Introduction}

For autonomous vehicles (AVs) to succeed as the future of automotive transportation, they must satisfy stringent safety requirements via rigorous verification and validation (V\&V) regimes. When demonstrating the safety of complex autonomous driving systems, it is a well-established standard to take a decompositional approach where constituent subsystems are evaluated in isolation \cite{iso2018road}. These results are then composed to create a holistic safety argument for the entirety of the system. Within AV development, a typical decomposition might entail separating the system into sensing, perception, behavior planning, and action components \cite{albus20024d}. In this work, we propose a novel method for defining and evaluating safety requirements that have been decomposed onto the \textit{obstacle detection} subsystems of AVs. 

Obstacle perception plays a critical role in providing an AV with awareness of obstacles (e.g., other vehicles, pedestrians) in its vicinity, and consequently have significant downstream effects on planning and control subsystems. Naturally, missing a detection or falsely reporting an obstacle may lead to collision or undesired behaviors.
While desired, it is impractical to expect an AV's obstacle perception module to provide high detection performance for all objects anywhere in the environment, particularly for visible but very distant obstacles. 
At the same time, to ensure safe driving operations, it is important to set a minimum threshold on the size of the region that the perception model is required to perform well in. Moving forward, we refer to such a region as a ``\textit{perception zone}''.
Naturally, an important question is: 

\textit{How should we define the shape and size of a perception zone?}

\begin{figure}
    \centering
    \includegraphics[width=0.45\textwidth]{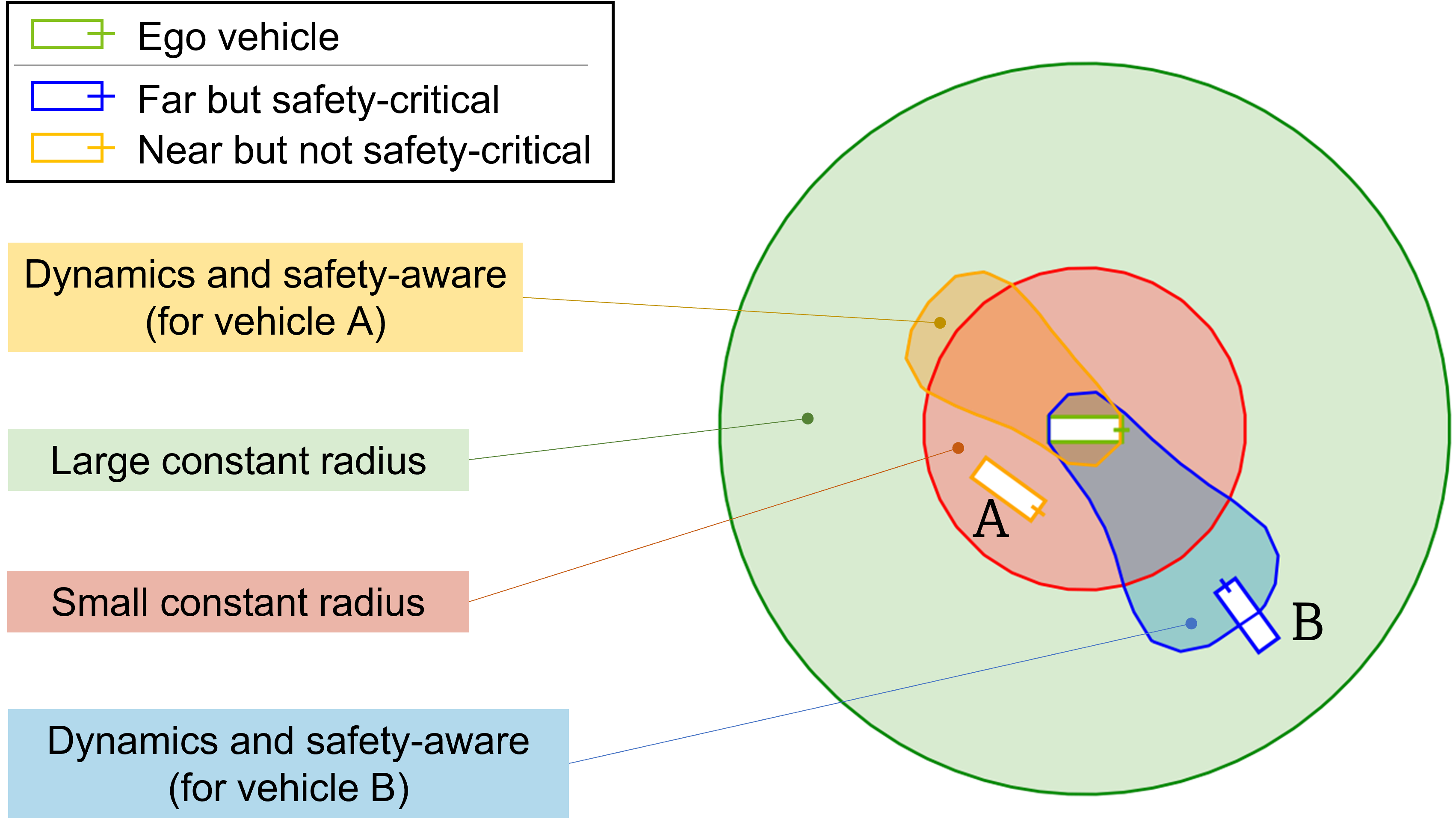}
    \caption{We propose an \textit{interaction-dynamics-aware} perception zone to inform an ego vehicle's perception module which obstacles are considered safety-critical. In contrast, na\"{i}ve approaches such as a small circular zone (red circle) may ignore distant obstacles that pose a safety threat to the ego vehicle (green vehicle), or an unnecessarily large circular zone (green circle) may impose overly-stringent requirements on the perception algorithm under test.}
    \label{fig:safety-aware percetion zone concept}
    \vspace{-0.5cm}
\end{figure}

At a conceptual level, it is well-understood that not all detection errors are equally damaging since not all obstacles are an imminent safety threat to the ego vehicle (i.e., not all obstacles are safety-critical). Therefore it is important to understand when a detection error involves a safety-critical obstacle, and when it does not.
The challenge is therefore (i) defining characteristics that delineate a practically-sized perception zone that encompasses any and all safety-critical obstacles, and (ii) designing tractable approaches to compute such a perception zone. 
Using a statically sized circular perception zone may be extremely computationally lightweight, but cannot adjust to the speed of the ego vehicle.
Prior work consider the dynamics of the vehicles/obstacles in computing such a perception zone but makes specific assumptions about how agents behave \cite{volk2020comprehensive,bansal2021risk} which may be poor reflections of what happens in practice. 

In this work, we put forth an approach to defining and computing perception zones that accounts for the \textit{dynamics} and \textit{closed-loop interaction behaviors} of agents. 
In particular, our proposed interaction-dynamics-aware perception zone, or ``\textit{safety zones}'' for short, accounts for a range of possible closed-loop policies by other agents, and captures notions that nearby vehicles moving away at a high speed from the slower-moving ego vehicle does not pose an imminent safety threat whereas a vehicle farther away but moving towards the ego vehicle may (see Figure~\ref{fig:safety-aware percetion zone concept}).
To account for the closed-loop interaction dynamics between agents when defining a safety zone, we borrow techniques from optimal control theory, namely Hamilton-Jacobi (HJ) reachability, \cite{mitchell2005time,margellos2011hamilton,bansal2017hamilton}, for which there are computationally tractable methods we can leverage.
We believe that perception zones that takes into consideration the interaction dynamics between agents can provide a more realistic depiction of which obstacles should be considered safety-critical, allowing for the implementation of better performance evaluators, and altogether increasing performance, safety, and trust in AVs.

\textbf{Statement of Contributions:} 
In summary, this work advances methods for defining regions of state space in which accurate perception is critical for safe operation of an AV system.
In particular, our contributions are three-fold: We \textbf{(i)} propose the use of closed-loop interaction dynamics, in addition to standard physical dynamics (e.g., vehicle dynamics), in defining a perception zone, \textbf{(ii)} demonstrate that our proposed safety zone can be expressed through the lens of HJ reachability for which there are existing computational tools which we can leverage. Through the HJ reachability framework, we show that the results are mathematically complete, and are easily adaptable to reflect a variety of safety requirements, and \textbf{(iii)} demonstrate, using an off-the-shelf perception algorithm from the nuScenes detection challenge leaderboard \cite{nuscenes}, that our approach is computationally lightweight to run, differs significantly from a baseline with weaker correctness properties, and better utilizes dynamic information.

\textbf{Organization:} Next, we describe current approaches to evaluating perception algorithms and techniques for identifying safety-critical perception errors in Section~\ref{sec:related_work}. Then in Section~\ref{sec:problem-formulation}, we highlight the importance of interaction dynamics in ensuring a perception zone is complete. Thereafter, we introduce HJ reachability in Section~\ref{sec:hj_reachability} and describe how we can use the HJ reachability framework to compute safety zones. We illustrate the impact of using our safety zones in Section~\ref{sec:results} via experiments on a nuScenes detection task. Finally, we discuss its usage beyond the scope of this paper and avenues for future work in Section~\ref{sec:discussion}, and conclude in Section~\ref{sec:conclusion}.

\section{Related Work} \label{sec:related_work}
A key challenge in evaluating and validating perception systems for safe automated driving tasks is understanding when perception errors are safety-critical. 
The exploration of this field is in its nascent stages. In this section, we discuss some existing approaches to tackling this problem by (i) defining task-aware (e.g., safety-aware) evaluation metrics, and (ii) defining what is considered ``safety-critical''. We note that there are parallel efforts in developing collision threat metrics to evaluate the safety of a \textit{planning} module (see \cite{dahl2019collision} for a summary). However, we stress that for safety considerations, perception evaluation should be decoupled from planning evaluation, and that perception requirements fundamentally differ from those of planning.
Furthermore, we note that obstacle occlusion is out of the scope of this work. We expect this to be covered by a separate perception signal such as a free-space region.

\subsection{Task-aware evaluation metrics}
\label{subsec:related:task-aware eval metrics}
Standard perception evaluation metrics such as Intersection over Union (IoU) and False Positive (FP) rates are ubiquitous, largely due to the fact that they are task-agnostic and provide good comparability across a variety of benchmarks. However, such task-agnostic metrics do not adequately quantify how well a perception model will actually perform when integrated into a full autonomy stack and deployed into the real-world. This is because the type of misdetection can lead to very different behaviors in downstream tasks \cite{aravantinos2020making}. 
The misalignment between typical perception metrics and application makes it challenging to rely on existing metrics to verify and validate perception algorithms deployed in the real-world.
In fact, using a case study on a pedestrian detection algorithm on an AV, \cite{lyssenko2021evaluation} showed there was a linear degradation in IoU performance the further away a pedestrian was, therefore indicating that IoU alone is not sufficient in validating safety. As such, they propose two different evaluation metrics that require user-specified inputs. Namely, (i) given a fixed distance between the AV and pedestrian, what is the smallest IoU of any pedestrian up to that distance, and (ii) given a minimum IoU value $\delta$, what is the maximum distance between the AV and pedestrian where $\delta$ is a lower bound on IoU. They argue that metrics such as these are more useful in determining if a perception algorithm is acceptable for the task. However, the question of selecting suitable thresholds remains.

More recently, there has been a push to use task-aware metrics as a gold standard when evaluating perception algorithms (see \cite{hoss2021review} for an overview). Specifically, the planning-aware metric proposed in \cite{philion2020learning} has been incorporated into the nuScenes detection task \cite{Guo2020efficacy}, one of the largest AV detection challenges available. The proposed planning-aware metric uses KL-divergence to compare how different the ego-vehicle's plans are with noisy and with perfect detection. Unsurprisingly, \cite{philion2020learning} showed empirically that the importance of detecting an object correctly depended on the distance from the ego vehicle and where the ego vehicle is likely to travel in the future, and they showed that their results were consistent with humans' intuition about which perceptual uncertainties are actually dangerous.
Concurrent to these efforts, there have been other proposed safety-aware metrics such as \cite{volk2020comprehensive} which proposes (linearly) combining scores measuring detection quality, collision potential, and time needed to make the detection. Another example is \cite{bansal2021risk}, which ranks objects based on their collision risk (imminent, potentially, none) defined via a simplified forward reachable set computation under an isotropic force assumption. 
While such task-relevant metrics are primarily useful in comparing the \textit{relative} performance of a perception algorithm over another, unfortunately, the metrics themselves are not useful in \textit{validating} whether a perception algorithm is sufficient in supporting safe AV operations.

\subsection{Safety-critical perception errors}
\label{subsec:related:safety-critical}

In general, safety validation of AVs requires demonstrating that the system can operate within an acceptable risk level specified by the appropriate regulatory body or industry safety standard.
Given this requirement, \cite{berk2020assessing} defines the safety-critical perception error rate as $\lambda_\mathrm{per}^\mathrm{safety\,crit.} = \lambda_\mathrm{per} \cdot p_\mathrm{per}$ where $\lambda_\mathrm{per}$ is the rate of perception errors and $p_\mathrm{per}$ is the fraction of safety-critical perception errors. While an intuitive definition, it is unclear and non-trivial how to compute a value for $p_\mathrm{per}$. \cite{berk2020assessing} suggests computing $p_\mathrm{per}$ empirically via closed-loop simulation (which is costly and leads to the broader challenge of simulation realism), or to define it heuristically.

As stated in \cite{berk2020assessing}, an unsafe perception error is one where a collision could have been prevented if the error had not occured. For example, consider the case where an AV fails to detect an obstacle. If the AV could have prevented the collision in the absence of the error, then a safety-critical perception error has occurred. \cite{salay2019safety} proposed a framework called classification failure mode effect analysis (CFMEA) to quantify the severity of a perception error but assumes access to a closed-loop policy evaluation mechanism (e.g., simulator), which may be costly to run for large scale simulations, and provide inaccurate results due to the non-trivial sim-to-real gap.
Unfortunately, precisely pinpointing safety-critical perception errors is challenging, not only because obtaining ground truth data is hard, but also due to the closed-loop nature of AV operations and temporal correlation in sensor readings. 

\begin{figure}
    \centering
    \includegraphics[width=0.45\textwidth]{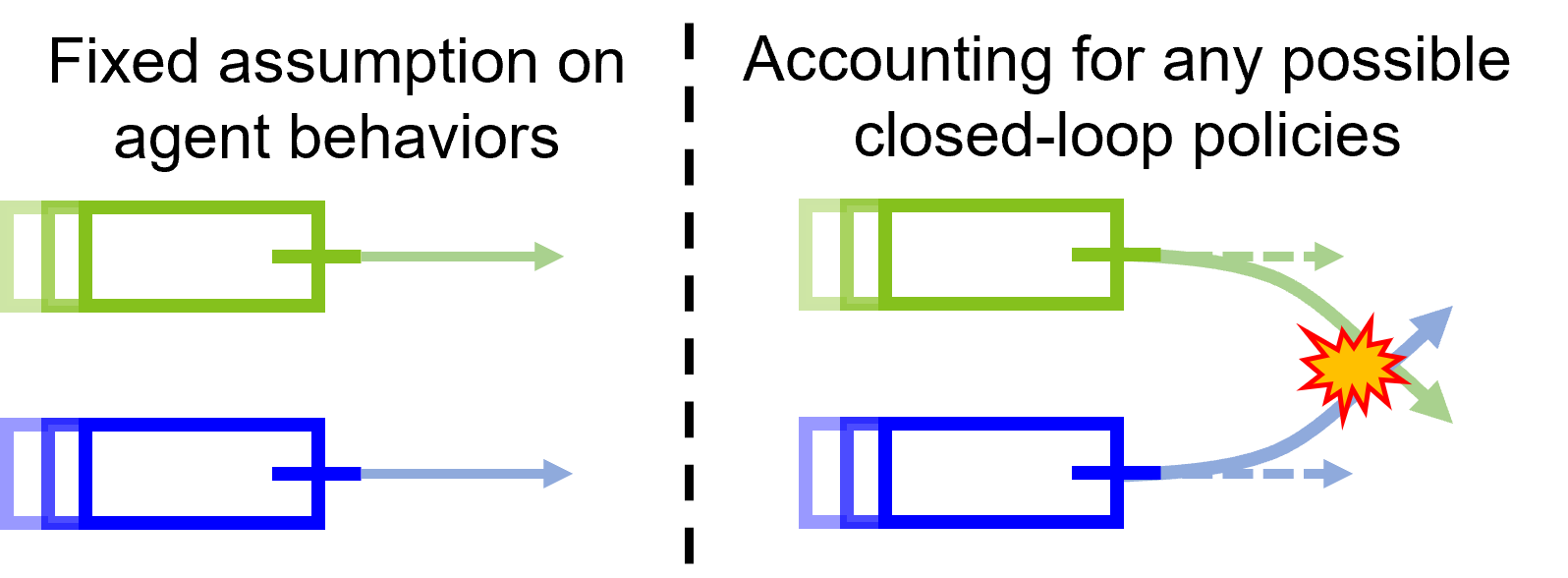}
    \caption{Overly-restrictive assumptions on agent behaviors such as braking only with no steering (left) ignores possible interactive behaviors which may lead to collisions (right). It is therefore important to account for such possible interactions when computing perception zones.}
    \label{fig:interaction_important}
    \vspace{-0.5cm}
\end{figure}

Instead, a more tractable approach is to determine which obstacles in the environment are safety-critical and ensure that perception performance is high for those obstacles. 
Along this vein, there has been a number of proposed approaches, many of which can be viewed as specific instantiations of a reachability problem \cite{bajcsy2021towards}. In \cite{volk2020comprehensive}, they leverage the responsibility-sensitive safety (RSS) framework \cite{shalev2017formal} which considers another vehicle as safety-critical if a collision with the ego-vehicle would occur if both vehicles were to brake and come to a stop.
Similarly, \cite{bansal2021risk} considers another vehicle to be safety-critical if a collision with the ego-vehicle occurs when both vehicles continue moving with constant velocity (i.e., zero control input).
While these types of safety assessment consider the agents' dynamics, they nonetheless make strong assumptions on agent behavior, including assumptions on the actions of \textit{other} vehicles (e.g., perform hard brake) which the ego-vehicle unfortunately does not have control over. Furthermore, they do not account for possible \textit{reactions} (i.e., interactions), between agents. Therefore, the safety assessment is invalid if the other vehicle behaves differently to what is assumed (see Figure \ref{fig:interaction_important}).

A more robust but more conservative approach is to account for all possible, including worst-case, closed-loop policies agents can take. HJ reachability \cite{mitchell2005time} is such a framework to model such interactions, but has been typically applied to safety-critical control settings \cite{bajcsy2019efficient,leung2020infusing} and not for perception applications.

\section{The Importance of Interaction Dynamics}\label{sec:problem-formulation}

The goal is to define and construct an ``appropriately-sized'' perception zone that encompasses safety-critical obstacles only. That is, we desire our zone to be sufficiently large to encompass all safety-critical obstacles, but not too small such that it omits any.
Concretely, given a definition of safety-criticality (such as the one introduced at the end of this section), two key desiderata for such a perception zone are soundness and completeness with respect to safety-criticality, as defined below.

\begin{definition}[Sound]
For all obstacles within a scenario, if an obstacle is within the perception zone, the obstacle is safety-critical.
\label{def:sound}
\end{definition}

\begin{definition}[Complete]
For all obstacles within a scenario, if an obstacle is safety-critical, the obstacle is within the perception zone.
\label{def:complete}
\end{definition}

As a bare minimum, a perception zone should capture all safety-critical obstacles. Therefore, completeness is a necessary condition, while soundness is not. Of course, both properties are ideally desired, but the soundness criterion can be relaxed to make constructing a perception zone computationally tractable. An unsound perception zone will simply impose higher stringency requirements on the perception module under test, since the zone will be larger than strictly necessary. In short, the less sound perception zones are, the more unnecessarily stringent the perception requirements become.

While relaxing the strict soundness requirement simplifies the problem, achieving completeness is still a difficult property to incorporate into a safety zone definition due to the closed-loop nature of agent behaviors. In Section~\ref{subsec:related:safety-critical}, we reviewed several approaches to defining safety-critical regions of state space. However, they all make strong assumptions about system behavior which may not always hold in practice, and are thus incomplete because they may classify an obstacle as not safety-critical even though it is possible for it to collide with the ego vehicle. Consider for example Figure~\ref{fig:interaction_important}: If we strictly assume both agents will maintain heading and brake (left figure), then they will not collide and the contender (blue/bottom) is not considered safety-critical. On the other hand, if we consider any possible closed-loop policies that each agent can take, such as maintaining heading or swerving into each other (right figure), then a collision is possible and therefore the contender is considered safety-critical.

To the best of the authors' knowledge, existing approaches fail to take into consideration \emph{interaction dynamics}, that is, closed-loop reactive control policies of each agent.
Factoring in these effects requires sophisticated modeling methods. Choosing na\"{i}ve dynamics models, such as circles derived under worst-case assumptions, may be complete but would be extremely unsound (i.e. unnecessarily large).

This motivates the need to design a perception zone that considers (i) reasonably accurate dynamics models, and (ii) all possible closed-loop interactions between agents.
Moreover, since the end goal is to use the perception zone within an industrial V\&V effort, there are additional practical considerations we like to keep in mind. The perception zone ideally should also be  computationally efficient, easy to adapt to different safety requirements, and ideally independent from planning and control module implementations such that a decompositional safety argument can be made
\cite{iso2018road,albus20024d}.

In this work, we demonstrate that a robust optimal control approach, namely, HJ reachability, is an attractive solution to compute such an interaction-dynamics-aware perception zone (a.k.a., safety zone). In subsequent sections we will describe our proposed HJ reachability solution in detail, and demonstrate its practicality with experimental results.
However, we first introduce a specific safety requirement and use this as a representative use-case throughout the paper.

\textbf{Representative use-case:}
We define an obstacle as safety-critical if it can collide into the ego vehicle at any location, speed, or severity. \footnote{In general, we can decompose collision further according to different levels of exposure or severity.}
Without loss of generality, we will focus on the following prototypical safety requirement for preventing collisions resulting from false positive (FP) \textit{vehicle} detections (i.e., detecting a vehicle is present when it is actually not there): 

\begin{quote}
    The perception subsystem shall not report false positive vehicle detections that may cause unintended ego-vehicle deceleration greater than 3.5 ms$^{-2}$ after a 0.5s reaction delay.
\end{quote}

In other words, if it is possible for the ego vehicle to collide with a falsely detected vehicle (i.e. a ghost obstacle) while limiting its deceleration, the target obstacle is considered safety-critical. The motivation for this safety requirement is to prevent unnecessary aggressive deceleration that may lead to rear-end collisions. Moving forward, our algorithmic design choices and experiments will be based on this representative use case, but we note that our proposed method is general and can capture alternative safety requirements.

\begin{figure*}[t]
    \centering
    \includegraphics[width=\textwidth]{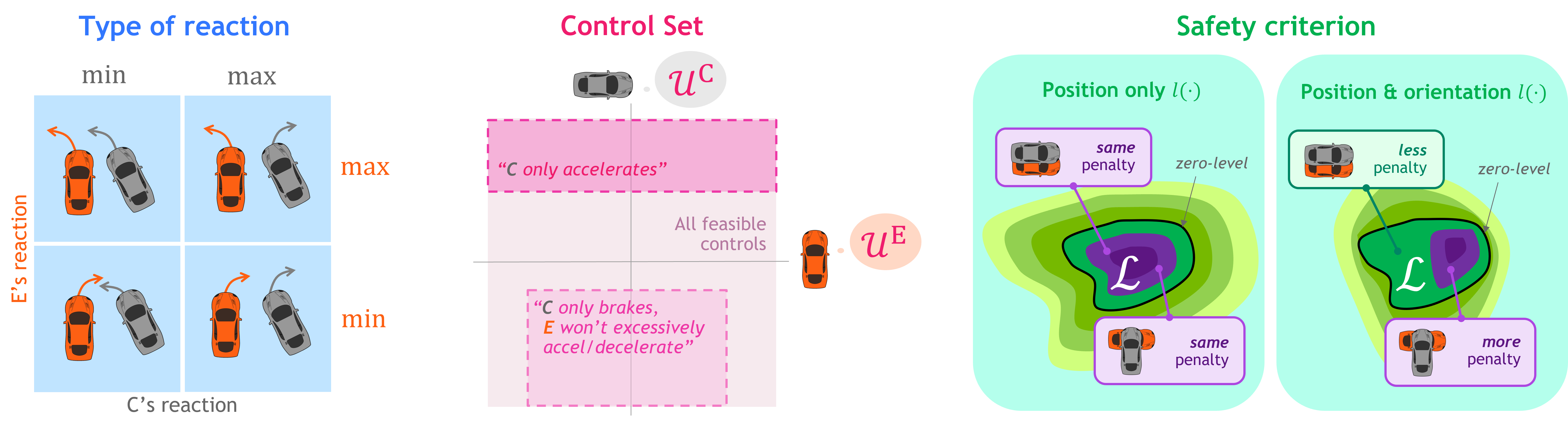}
    \caption{Elements of HJ reachability that a designer can adjust to reflect their design requirements.}
    \vspace{-1em}
    \label{fig:elements_of_hj}
\end{figure*}

\section{Computing Safety Zones via HJ Reachability}
\label{sec:hj_reachability}
In this section, we formally introduce HJ reachability and show that the safety zone can be constructed by solving an instantiation of a HJ reachability problem.
HJ reachability is a good fit for this task because (i) it is a flexible framework that can capture a variety of systems and specifications, and (ii) the results are cached offline allowing for fast and lightweight look-ups online. The main drawback is that the offline solution process suffers from the curse of dimensionality, but, as we discuss later, with some relaxations, we are able to keep our problem tractable.

\subsection{Hamilton-Jacobi reachability} %

HJ reachability is a mathematical formalism, advanced by the optimal control literature, for characterizing the performance and safety properties of (multi-agent) dynamical systems \cite{mitchell2005time,margellos2011hamilton,bansal2017hamilton}. Core to HJ reachability is the notion of a set of target states $\mathcal{L}$ which agents reason about either seeking or avoiding within a time horizon $T$. For example, $\mathcal{L}$ may correspond to the set of joint system states where the ego agent $E$ and contender $C$ are in collision (in general, $\mathcal{L}$ can represent different criteria depending on the application).
The output of a HJ reachability computation is a set of initial states (termed the backwards reachable tube, or BRT) where membership denotes that it is possible for agents to enter the set $\mathcal{L}$ at some point in the future (e.g., it is possible for the agents to collide) while following their respective optimal control policies.

Contrary to many applications of HJ reachability theory in safety-critical settings \cite{leung2020infusing,fisac2018general}, in this work we do not consider an adversarial setting where an ego agent $E$ is trying to avoid collision while subject to the worst-case (collision-seeking) actions of the contender agent $C$. Instead, in the context of possibly erroneous perception (specifically, false positive detections), we will make an even more conservative assumption that the ego agent $E$ may be in a situation where its preferred actions happen to be collision-seeking as well. That is, we consider a false positive detection to be \emph{safety-critical} if it is at all possible for the two agents to collide. Formally, given joint dynamics $\dot z = f(z, \uA, \uB)$, where $z \in \mathcal{Z} \subset \mathbb{R}^n$ denotes the joint state of agents $E$ and $C$, and $\uA$ and $\uB$ are the controls of agents $E$ and $C$ respectively, we define
\begin{equation}\label{eq:brt}
\begin{aligned}
\mathcal{S}(t) := &\left\{\bar z \in \mathcal{Z} \mid \exists \uA(\cdot), \exists \uB(\cdot), \exists s\in[t, 0],\right.\\
                 &\ \left. z(t) = \bar z \wedge \dot z = f(z, \uA, \uB) \wedge z(s) \in \mathcal{L}\right\}.
\end{aligned}
\end{equation}
In words, $\mathcal{S}(t)$ denotes a set of joint initial states where there exists a policy $\uA(\cdot)$ and $\uB(\cdot)$ where agents $E$ and $C$ may enter $\mathcal{L}$ (e.g., collide if $\mathcal{L}$ represents collision states) within a time horizon $|t|$ in the future.

The set $\mathcal{S}(t)$ may be computed as the zero sublevel set of a value function $V(z, t)$ which obeys a Hamilton-Jacobi-Bellman partial differential equation (PDE),
\begin{equation}
\begin{aligned}
    \frac{\partial V(z, t)}{\partial t} &+ 
    \min \{0,\mathcal{H}(z, t) \} = 0 , \qquad V(z, 0) = {\color{ForestGreen}\ell(z)},\\
    \mathcal{H}(z,t) &= {\begingroup
     {\color{brightBlue} \min}_{\uA \in {\color{WildStrawberry} \UA}} {\color{brightBlue} \min}_{\uB \in {\color{WildStrawberry}\UB}}\nabla_{z} V(z, t)^\top f(z, \uA, \uB),
    \endgroup}\\
\end{aligned}
\label{eq:hji_pde_brt}
\end{equation}

where the boundary condition for this PDE is defined by the function ${\color{ForestGreen}\ell}: \mathcal{Z} \rightarrow \mathbb{R}$ whose zero sublevel set encodes $\mathcal{L}$, i.e., $\mathcal{L} =  \lbrace z \mid {\color{ForestGreen}\ell(z)} < 0\rbrace$.
Critically, \eqref{eq:hji_pde_brt} accounts for \textit{closed-loop} control policies of both againts because at any point in time and at any state, \eqref{eq:hji_pde_brt} considers $\uA$ and $\uB$ that minimizes the value function.

By solving Equation~\eqref{eq:hji_pde_brt} \textit{backwards} in time over a time horizon of $T$, we obtain the HJ value function $V(z,t)$ for $t\in[-T,0]$. For any starting state $z \in \mathcal{Z}$, $V(z,t)$ corresponds to the lowest value of {\color{ForestGreen}$\ell(\cdot)$} along a system trajectory within $|t|$ seconds if both agents $E$ and $C$ act optimally, $\uA^*(z),\,\uB^*(z) = {\color{brightBlue}\argmin}_{\uA \in {\color{WildStrawberry}\UA}}{\color{brightBlue}\argmin}_{\uB \in {\color{WildStrawberry}\UB}} \nabla_{z} V(z, t)^\top f(z, \uA, \uB)$. Thus the set of states from which collision may be reached within $|t|$ seconds is $\mathcal{S}(t) = \lbrace z \mid V(z, t) < 0 \rbrace$.

In this work we use \cite{HJJAX2021} to numerically solve the HJB PDE; the solver stores the value function over an $n-$dimensional grid in state space where $z\in\mathcal{Z}\subset \mathbb{R}^n$. Since solving the HJB PDE suffers from the curse of dimensionality, it is only practical for relatively small systems (roughly $<7$ dimensions). However, once the solution is computed and cached offline, the value at any state and time can be efficiently found via a look-up and/or linear interpolation during online computation.

\textbf{Summary:} By approaching the safety zone problem through the lens of reachability theory, we are able to compute the set of joint states $\mathcal{S}(t)$ where membership of the set indicates that entry into $\mathcal{L}$ (i.e., the safety requirement is violated) is possible within $|t|$ seconds when considering a set of possible closed-loop control policies that each agent can take. 
To reiterate, it is important to reduce the number of false positive detections a perception algorithm makes inside the safety zone because contenders inside the zone have the potential to violate the ego vehicle's safety requirement. False positive detection errors that lay outside the safety zone are considered harmless because a safety violation is not possible regardless.

\subsection{Adjustable parameters of the HJ reachability problem}
\label{subsec:adjustable_params}
The HJ reachability problem setup described by \eqref{eq:hji_pde_brt} is only one possible formulation. As indicated by the colored text in \eqref{eq:hji_pde_brt}, the formulation has a few elements, or ``knobs'', that a user can manipulate to reflect a variety of assumptions they may want to enforce.

\noindent{\color{brightBlue} \textbf{Type of agents' reactions ($\max \text{ or } \min$)}}: 
By modifying the information pattern and/or if agents are minimizers or maximizers, we can modify the strategic optimism when computing the unsafe set (see Figure~\ref{fig:elements_of_hj}, left). In this work we consider a $(\min, \min)$ formulation while, as previously noted, another common approach is to assume agent $E$ is collision-avoiding (instead of collision-seeking) which corresponds to a $(\max, \min)$ formulation. In the context of perception zones, the $(\min, \min)$ formulation enables us to make completeness arguments (see Section~\ref{subsec:completeness}).

\noindent{\color{WildStrawberry} \textbf{Allowable control sets ($\UA$ and $\UB$)}}: Traditionally, the control set represents all dynamically feasible controls of each agent (e.g., respecting actuation limits of the system). However, we can \textit{restrict} the control set to reflect assumptions about how other agents behave in safety-critical scenarios (see Figure~\ref{fig:elements_of_hj}, center). In this work, we consider the ego vehicle coming to a stop while freely steering after a reaction delay, and therefore we restrict its acceleration capabilities accordingly (see Section~\ref{subsec:assumptions_on_behaviors}). This ego vehicle behavior assumption stems from our definition of safety-criticality.

\noindent{\color{ForestGreen} \textbf{Safety criterion $\mathcal{L} = \lbrace z \mid \ell(z) < 0\rbrace$}}: We are free to define {\color{ForestGreen}$\ell(\cdot)$} as long as its zero sublevel set equals $\mathcal{L}$.
While purely geometric functions like penetration/separation distance are a common choice, we can design or learn alternative functions that capture more nuanced notions of safety. For example, by shaping {\color{ForestGreen}$\ell$} to penalize more dangerous orientations (e.g., T-bone or head-on collision) we can encode collision severity or collision responsibility (see Figure~\ref{fig:elements_of_hj}, right). As we will describe in Section~\ref{subsec:assumptions_on_behaviors}, we can use the solution of one HJ reachability problem as the initial condition for a second HJ reachability problem to model interactions with multiple phases.

\subsection{Vehicle dynamics modeling}
An important design choice is the selection of the dynamics model $f(\cdot, \cdot, \cdot)$. A higher fidelity model can better represent the true system but can make the HJ reachability computation quickly intractable as it suffers from the curse of dimensionality.
To address this trade-off, we relax the safety zone soundness requirement as described in Section \ref{sec:problem-formulation} while maintaining completeness. Essentially, we elect to use a lower fidelity dynamics model (i.e., ignore higher order derivatives such as jerk and steering rate)
to keep the HJ computation tractable at the cost of being slightly over conservative (i.e., no longer sound); see Section~\ref{subsec:completeness} for additional discussion.

The dynamics model considered will depend on the type of obstacle detected (e.g., vehicle, pedestrian, cyclist). For this discussion, we restrict our exposition to \textit{vehicles} and note that similar computations can be performed under different dynamics.
We assume the ego and the contender vehicle each obey the dynamically extended simple car model described by the following dynamics,
\begin{equation}
    \dot{w} = \begin{bmatrix}
    \dot{x} \\ \dot{y} \\ \dot{\psi} \\ \dot{v}
    \end{bmatrix} = \begin{bmatrix}
    v\cos\psi\\ v\sin\psi \\ \frac{v}{d}\tan\delta\\ a
    \end{bmatrix}, \quad \begin{matrix}
    w = [x, y,\psi, v]^T & \text{(state)}\\
    u = [\delta, a]^T & \text{(controls)}\\
    0 \leq v \leq v_{\max}&\\
    u_{\min} \leq u \leq u_{\max}&
    \end{matrix}
    \label{eq:dynamically extended simple car}
\end{equation}
where $(x, y)$ is the position of the center of the rear axle in a fixed world frame, $\psi$ is the heading angle, $v$ is the speed, $\delta$ is the steering input, $a$ is the acceleration input, and $d$ is the distance between the front and rear axle.
Let states, controls, and parameters with subscripts $\mathrm{E}$ and $\mathrm{C}$ denote the ego and contender vehicle respectively. 
Define the relative state as $z=[\xrel, \yrel, \psirel, \vA, \vB]$ and associated dynamics as,

\vspace{-6mm}
{\small 
\begin{equation}
    \dot{z} = \begin{bmatrix}
        \vB \cos\psirel -\vA + \frac{\yrel  \vA }{\LA}\tan\deltaA\\
        \vB \sin\psirel - \frac{\xrel \vA }{\LA}\tan\deltaA\\
         \frac{\vB}{\LB}\tan\deltaB -  \frac{\vA}{\LA} \tan\deltaA\\
        \aA\\
        \aB
    \end{bmatrix}, \,
    \begin{matrix}
    \begin{bmatrix}
    \xrel\\ \yrel
    \end{bmatrix} = \mathrm{R}_{\psiA}\begin{bmatrix} \xB - \xA \\ \yB - \yA \end{bmatrix},\\
    \\
    \mathrm{R}_{\psi} = \begin{bmatrix} \cos \psi & \sin \psi\\ -\sin\psi & \cos\psi \end{bmatrix},\\
    \\
    \psirel = \psiB - \psiA.
    \end{matrix}
    \label{eq:relative dynamics}
\end{equation}
}
\!\!We use the relative dynamics in \eqref{eq:relative dynamics} when solving the PDE~\eqref{eq:hji_pde_brt}.

\subsection{Terminal value function}
The terminal value function $\ell(\cdot)$ can be designed to reflect different safety requirements such as front or rear end collisions within a pre-specified velocity range. For simplicity, we consider any type of collision at any speed as unsafe. Therefore we define $\ell(\cdot):= \ell_\mathrm{SD}(\cdot)$ where $\ell_\mathrm{SD}(\cdot)$ describes the signed distance between the two vehicles modeled as two rectangular rigid objects.

{\small
\begin{table}[]
    \centering
    \begin{tabular}{|C{0.9cm}||C{1cm}C{1.5cm}C{1.cm}C{2.2cm}|}
    \hline
        \textbf{Phase} & \textbf{Duration [s]} & \textbf{Ego control} & \textbf{Contender control}& \textbf{Initial value $\ell(\cdot)$} \\ \hline\hline
        Reaction & $\displaystyle\Delta t_\mathrm{react}$ & Any & Any & $\displaystyle V_\mathrm{brake}(\cdot, -t_\mathrm{stop}^\mathrm{E}(\cdot))$ \\ \hline 
        Braking & $\displaystyle\frac{\vA}{a_\mathrm{brake}}$ & Fixed decel., any steering & Any & $\displaystyle \ell_\mathrm{SD}(\cdot)$\\ \hline
    \end{tabular}
    \caption{HJ reachability set-up for the reaction and braking phases.}
    \label{tab:vehicle behavior assumptions}
\end{table}
}

\subsection{Assumptions on agent behavior}
\label{subsec:assumptions_on_behaviors}
In the case of a safety-critical setting, a common approach is to assume the AV will default to a prespecified not-at-fault behavior \cite{shalev2017formal,vaskov2019not}. Any collision after the AV has come to a stop is not considered at fault.
In this work, we assume the ego vehicle will perform a hard braking maneuver and come to a complete stop. Furthermore, we allow for a reaction time $\Delta t_\mathrm{react}$ before the ego starts to brake at a fixed deceleration $a_\mathrm{brake}$.
The reaction time allows for any possible latency that the ego vehicle may experience before the braking is initiated.
To compute the HJ value function accounting for the two phases, noting that we are computing backward in time, we first compute the HJ value function solution for the braking phase $V_\mathrm{brake}(z,t)$ by using $\ell_\mathrm{SD}(\cdot)$ as the terminal value function. Then for the reaction phase, we set $\ell_\mathrm{react}(z) = V_\mathrm{brake}(z,-t_\mathrm{stop}^\mathrm{E}(z))$ as the terminal value function where $t_\mathrm{stop}^\mathrm{E}(z=[\xrel, \yrel, \psirel, \vA, \vB])=\frac{\vA}{|a_\mathrm{brake}|}$ is the time taken for the ego vehicle to come to a complete stop when applying a constant deceleration $a_\mathrm{brake}$.
The parameters of the HJ reachability computation for each phase are summarized in Table~\ref{tab:vehicle behavior assumptions}.

Since we take into consideration the stopping time in the braking phase, and the reaction time is over a fixed time interval, we can simply consider the last time slice from the resulting value function.

\subsection{Completeness of safety zone}\label{subsec:completeness}
In this subsection we formalize three aspects of the ``conservatism'' of our approach regarding (i) reduced model fidelity, and (ii) the application of our pairwise agent analysis to $n$-agent settings, and (iii) mobility limitations imposed by road boundaries and obstacles.
\subsubsection{Model fidelity}\label{subsubsec:model_fidelity}
The takeaway here is that by using a lower fidelity model compared to a higher fidelity model with higher-order integrator states (e.g., jerk, steering rate), we are only more conservative in our safety zone computation and therefore completeness is still preserved, but we relax our soundness requirement.

Consider extended dynamics
\[
\footnotesize
\dot z_\mathrm{ext} = f_\mathrm{ext}\left(z_\mathrm{ext}=\begin{bmatrix}
z\\
\uA\\
\uB\\
\dot u_\mathrm{A}\\
\dot u_\mathrm{B}\\
\vdots\\
\frac{\mathrm{d}^{i-1}\uA}{\mathrm{dt}^{i-1}}\\
\frac{\mathrm{d}^{j-1}\uA}{\mathrm{dt}^{j-1}}
\end{bmatrix}, {\textstyle\frac{\mathrm{d}^i\uA}{\mathrm{dt}^i}}, {\textstyle\frac{\mathrm{d}^j\uB}{\mathrm{dt}^j}}\right) = \begin{bmatrix}
f(z, \uA, \uB)\\
\dot u_\mathrm{A}\\
\dot u_\mathrm{B}\\
\ddot u_\mathrm{A}\\
\ddot u_\mathrm{B}\\
\vdots\\
\frac{\mathrm{d}^{i}\uA}{\mathrm{dt}^{i}}\\
\frac{\mathrm{d}^{j}\uA}{\mathrm{dt}^{j}}
\end{bmatrix}
\]
with integrators added to the controls to reflect realistic slew rates (as opposed to instantaneous changes) in, e.g., acceleration and steering. We claim that the safety zone $\mathcal{S}_\mathrm{ext}$ corresponding to these extended dynamics is a subset of $\mathcal{S}$, the safety zone corresponding to the lower fidelity dynamics $f$ (and thus completeness is still preserved by computing with the lower fidelity dynamics). To see this, we refer to the definition~\eqref{eq:brt} of the safety-critical set. If there exist controls $\frac{\mathrm{d}^i\uA}{\mathrm{dt}^i}(\cdot), \frac{\mathrm{d}^j\uB}{\mathrm{dt}^j}(\cdot)$ in the higher fidelity dynamics that lead to collision, then we may simply consider the corresponding lower fidelity controls $\uA(\cdot), \uB(\cdot)$ integrated from these higher fidelity controls as an existence proof for membership in $S$. That is, ignoring slew-rates (or analogously, other higher-fidelity modeling considerations) can only increase the control authority of the agents to seek collision with each other, and thus the set of safety critical states can only grow.

\begin{figure*}[t]
    \centering
    \includegraphics[width=\textwidth]{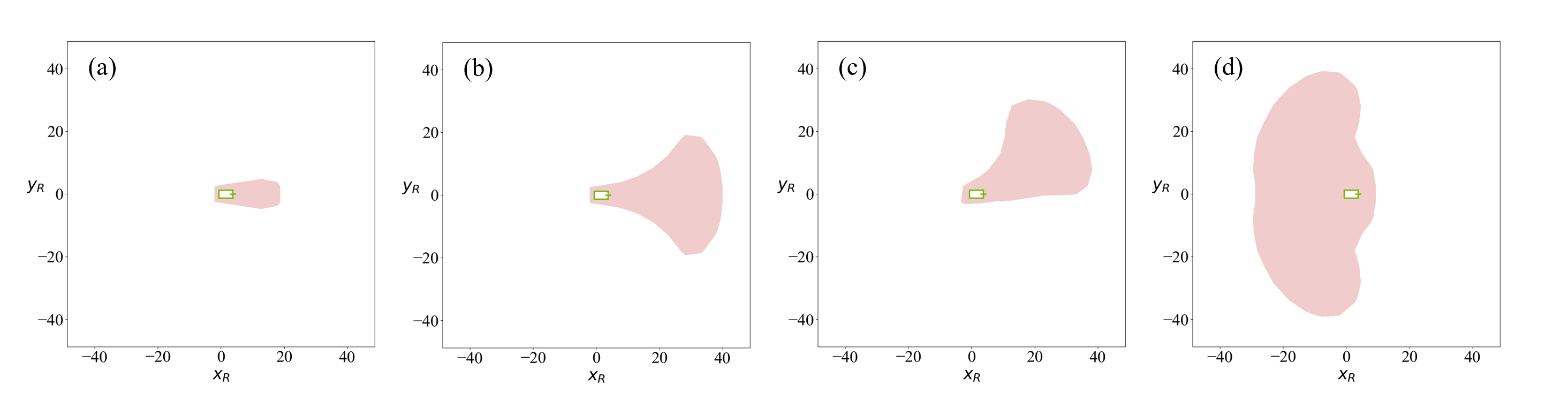}
    \caption{Visualizations of the safety zone for FP detections calculated using HJ reachability. We show the $\xrel$-$\yrel$ slice of state space taken at different relative angles $\psirel$, ego velocity $\vA$ and contender velocity $\vB$. The green rectangle represents the ego vehicle. The time horizon $t$ is always taken to be the stopping time (including reaction delay) at the given ego velocity. \textbf{(a)} $\vA = 5$ms$^{-1}$, $\vB = 5$ms$^{-1}$ and $\psirel = -\pi$. \textbf{(b)} $\vA = 7.5$ms$^{-1}$, $\vB = 7.5$ms$^{-1}$ and $\psirel = -\pi$. \textbf{(c)} $\vA = 7.5$ms$^{-1}$, $\vB = 7.5$ms$^{-1}$ and $\psirel = -\frac{3\pi}{4}$. \textbf{(d)} $\vA = 9$ms$^{-1}$, $\vB = 10$ms$^{-1}$ and $\psirel = 0$.}
    \label{fig:zone-examples}
\end{figure*}

\begin{figure}[t]
    \centering
    \includegraphics[width=\columnwidth]{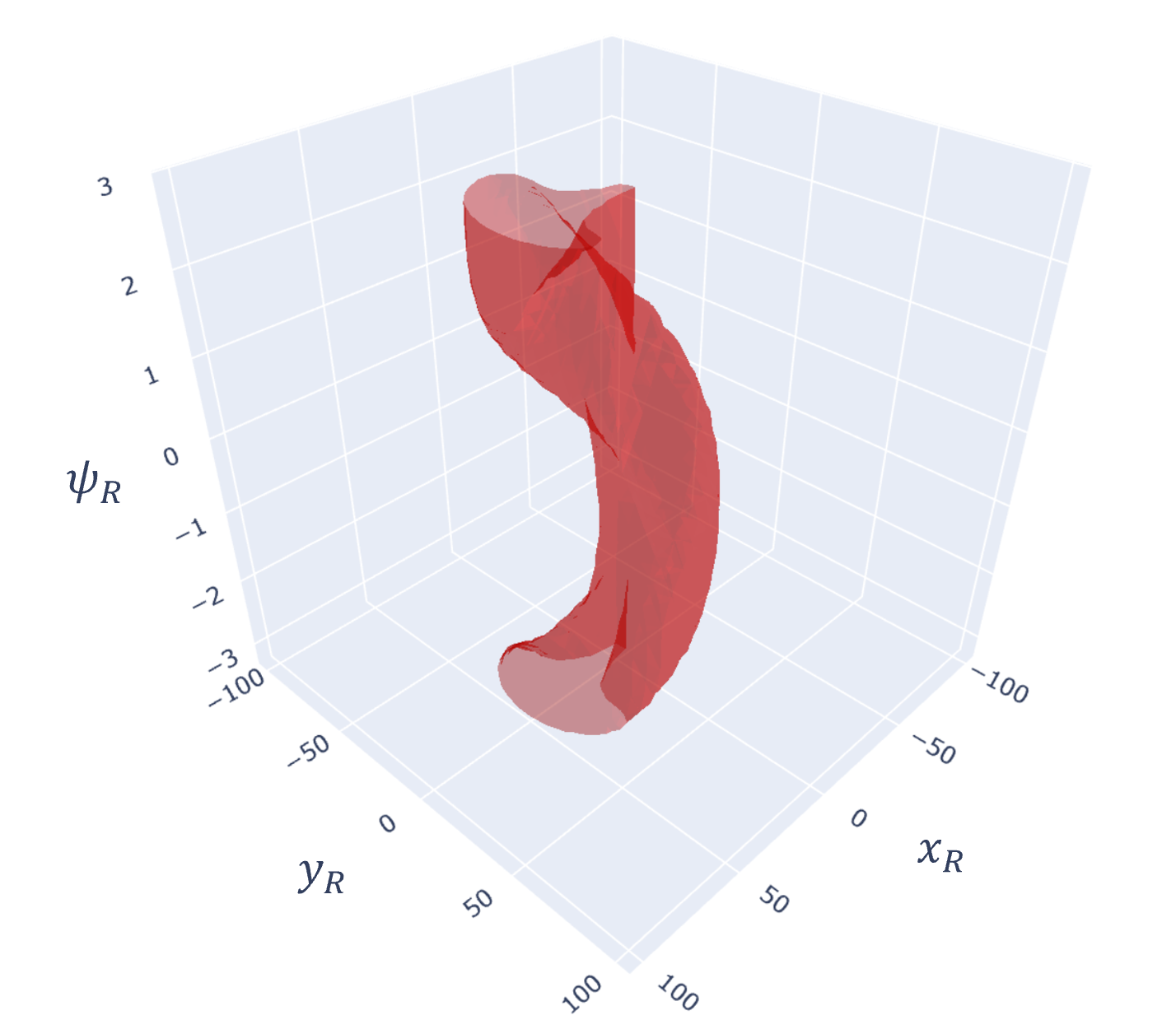}
    \caption{A 3-dimensional plot of our proposed safety zone for FP detections. We show the zone as a surface taken at the $\xrel$-$\yrel$-$\psirel$ slice of state space, with ego velocity $\vA$ and contender velocity $\vB$ both set to 6.5ms$^{-1}$, and time $t$ set to the stopping time of the ego vehicle (including reaction delay).}
    \label{fig:3d-plot}
    \vspace{-0.5cm}
\end{figure}

\subsubsection{Multi-agent extensions}
So far, we have considered agents only pairwise, with an ego agent $E$ interacting with a single contender $C$. One might also imagine an $n$-agent setting, where the ego agent $E$ could reason about joint dynamics $f_\mathrm{multi}(z_\mathrm{multi}, \uA, \dots)$ with multiple other agents. Similar to Remark~\ref{subsubsec:model_fidelity}, we argue that considering agents pairwise still preserves completeness, as long as the existence of multiple agents does not increase the control authority of any individual contender (indeed, conversely you would expect that contenders might interfere with/block each other off from colliding with the ego). Specifically, given a safety-critical joint state $z_\mathrm{multi} \in \mathcal{S}_\mathrm{multi}$ (i.e., there exist joint controls such that at least one contender $K$ collides with the ego) we can consider the restriction of the joint controls to just those of $E$ and $K$. This multi-agent configuration would thus still be regarded as safety-critical for the $(E, K)$ pair, i.e., it is sufficient to consider the safety-criticality of each false positive detection individually.

\subsubsection{Mobility limitations imposed by road boundaries and obstacles}
So far in our safety zone computation, we are considering only pairwise interactions and assume that everywhere is feasible driveable space. But in reality, there are other obstacles and road boundaries that would limit the mobility of the ego and contender vehicle. For instance, there could be a physical road boundary or other obstacles that a contender vehicle cannot physically pass through to reach the ego vehicle. However, ignoring the limited mobility issues imposed by road boundaries and other obstacles only make our approach more conservative and preserves completeness. Contrary to decision making, the added conservativeness is not critical for evaluating the perception system. Moreover, it is non-trivial to account for mobility limitations imposed by road boundaries and other obstacles. The added problem complexity could make HJ reachability computationally burdensome, and accurate knowledge of road boundaries and obstacles would itself rely on a perception system that needs validation, or expensive ground truth labeling. We plan to investigate tractable means to incorporate road geometry into the HJ computation to reduce the conservativeness.

\section{Results}
\label{sec:results}
With our methodology in place, we present our resulting interaction-dynamics-aware perception zone, and perform an empirical comparison between our approach and a baseline method on the nuScenes dataset \cite{nuscenes}.

\subsection{Safety-aware perception zone}

In order to perform the HJ reachability computation, we leverage the implementation found in \cite{HJJAX2021}. The resulting zones can be seen in Figures~\ref{fig:zone-examples} and \ref{fig:3d-plot}. 
Notice that the shape and size of slices of the safety zone vary in non-trivial ways as the agents' velocity and relative heading varies. This is attributed to that fact that the HJ computation takes into consideration the dynamics and closed-loop interaction between agents.

\begin{table}[h]
\centering
\begin{tabular}{|c|c|}
\hline
\textbf{Parameter} & \textbf{Value}\\ \hline\hline
Detection score threshold    & $0.3$                  \\
IoU threshold                & $0.5$                  \\
Maximum velocity $v_\mathrm{max}$ & $20\mathrm{ms}^{-1}$ \\ 
Ego deceleration         & $-3.5\mathrm{ms}^{-2}$            \\
Contender acceleration limit & $[-4.5\mathrm{ms}^{-2},4.5{ms}^{-2}]$ \\
Vehicle steering limit     & $[-10^\circ,10^\circ]$ \\
Vehicle size $[L\times W]$                 & $4.5\mathrm{m}\times2.5\mathrm{m}$       \\
Front-rear axle distance $d$ & 3m\\
Reaction time $\displaystyle\Delta t_\mathrm{react}$ & 0.5s\\ \hline
\end{tabular}
\caption{Parameter values used in our safety zone computation and experiments.}
\label{tab:params}
\end{table}

A list of the parameters used in our model can be found in Table~\ref{tab:params}. We discretized the state space with a grid of size $ 40 \times 40 \times 20 \times 15 \times 15$ for states $[\xrel, \yrel, \psirel, \vA, \vB]$ respectively. The high accuracy solver setting was used to simulate dynamics for both braking and reaction phases.

\subsection{Perception evaluation with interaction-dynamics-aware perception zones}

\begin{figure*}[t]
    \centering
    \includegraphics[width=1\textwidth]{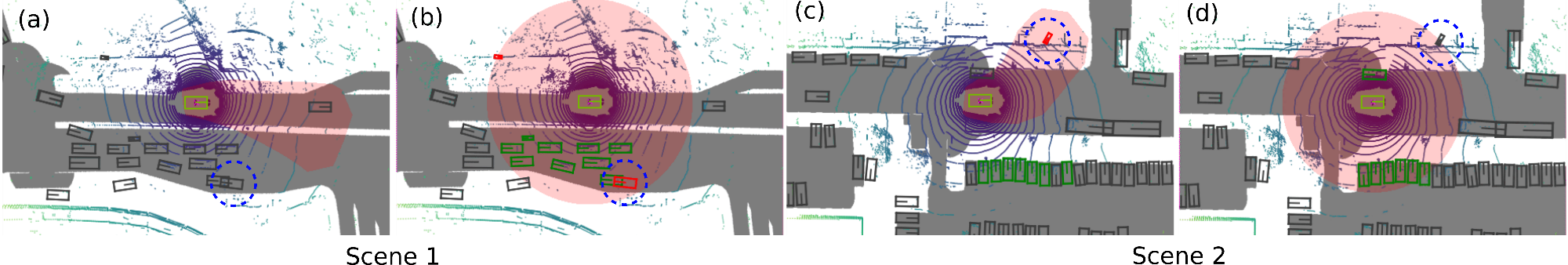}
    \caption{Comparison between our safety zone and the circular baseline. Slices of our proposed safety zone (red region) are taken with respect to the ego vehicle and blue circled vehicle's relative state, which is a false positive detection. (a) HJ safety zone slice plotted with blue circled vehicle state: $v=0$ms$^{-1}$, $\psi=162^\circ$. Our proposed safety zone does not consider the false positive detection as safety-critical whereas the circular perception zone (b) does. (c) HJ safety zone slice plotted with the blue circled vehicle state: $v=0\text{ms}^{-1},\psi=-119^\circ$. Our proposed safety zone considers the false positive detection as safety-critical whereas the circular perception zone (d) does not. Legend: {\color{ForestGreen}Green: safety-critical true positive vehicles}, {\color{darkgrey}Grey: non-safety-critical vehicles}, {\color{red}Red: safety-critical false positive}.}
    \label{fig:snapshot}
\end{figure*}
To illustrate the effect of the proposed method, we evaluate the nuScenes \cite{nuscenes} detection challenge result with \algname{}. We used the detection algorithm proposed in \cite{zhu2019class} as the baseline algorithm to generate bounding boxes for objects detected in the nuScenes detection dataset. The detected objects are then compared to the ground truth to determine whether it is an actual object or false positive. Moreover, by comparing subsequent frames, dynamic states such as the ego vehicle velocity and heading, and contender vehicle velocity and heading are obtained. With the dynamic states known, a slice of the HJ value function in position space is obtained (i.e., fixing $\psirel, \vA,$ and $\vB$ in the HJ value function). The safety-aware perception zone can be visualized in position space corresponding to regions where the HJ value is negative (see Figure~\ref{fig:snapshot}). 
Our proposed safety zone derived from HJ reachablity is compared with a baseline method where the perception zone is a circular region with a ego velocity-dependent radius,
\begin{equation*}
    r_\mathrm{stop}(\vA)=\vA\Delta t_\mathrm{react}+\frac{\vA^2}{2|a_\mathrm{brake}|}+\sqrt{L^2+R^2} ,
\end{equation*}
which is the stopping distance for the ego vehicle performing a constant deceleration $a_\mathrm{brake}$ after a reaction time $\Delta t_\mathrm{react}$ plus an added term accounting for the vehicle size, where $L,W$ are the assumed vehicle length and width. See Table~\ref{tab:params} for parameter values.

Figure~\ref{fig:snapshot} shows the comparison of our proposed safety zone and a baseline method; the red shaded region denotes the corresponding perception zones. Note that our safety zone depends on a selected contender's heading and velocity. Plots (a) and (c) show the safety zone corresponding to the contender's (circled in blue)  heading and velocity. Plots (b) and (d) show the baseline circular perception zone. In scene 1 (left), a contender that is a false positive detection (circled in blue) is stationary, positioned with some lateral offset, and facing in the opposite direction to the ego vehicle. Given the initial condition, even if the (falsely detected) contender tried its hardest to collide into the ego vehicle (i.e., accelerating and steering towards the ego vehicle) and the ego vehicle applied the worst-case steering (towards the contender), collision is not possible before the ego vehicle comes to a complete stop. Note that the presence of other obstacles and road boundaries would only make potential collision with the ego vehicle even less likely. Therefore ignoring other obstacles and road boundaries does not change the completeness property of the safety zone. As such, the proposed safety zone identifies the false positive detection in (a) as non-safety-critical. In contrast, the circular baseline method in (b) does not account for such dynamic interaction, and unnecessarily flags the false positive detection as safety critical.
Similarly, in scene 2 (right), a contender that is a false positive detection (circled in blue) is stationary but is headed directly towards the ego vehicle. Again, ignoring the presence of other obstacles and road boundaries, it is possible for a collision to occur given the relative heading and position. As such,  the proposed safety zone in (c) is able to flag the false positive detection as safety-critical whereas the baseline method (d) does not.

\begin{table}[]
\centering
\begin{tabular}{|c||c|c|c|}
\hline
 & FP count & FP rate & FP per frame \\ \hline \hline
Total & 4644 & 16.0\%  & 2.59 \\\hline
Safety-critical w/ HJ & 35 & 0.75\% & 0.013\\\hline
Safety-critical w/ circular & 305 & 6.6\%  & 0.113 \\ \hline        
\end{tabular}
\caption{Sample statistics comparing our proposed safety-aware perception zone with a baseline approach}\label{tab:stat}
\vspace{0.5cm}
\begin{tabular}{|c||c|c|}
\hline
 \backslashbox{HJ}{Circular} & Safety-critical  & Not safety-critical  \\ \hline \hline
Safety-critical&  22  & 13 \\
Not safety-critical  & 305 & 4326\\ \hline                                      
\end{tabular}

\caption{Cross comparison between our safety-aware perception zone and baseline circular zone.}\label{tab:comparison}
\end{table}
Table~\ref{tab:stat} shows an example of perception statistics using different perception zones, and Table~\ref{tab:comparison} shows the cross comparison between the two perception zone results. Comparing to the baseline circular perception zone, the HJ safety zone registers many fewer safety-critical false positive objects by leveraging extra dynamic information. However, it is not strictly less conservative than the baseline as there are cases where the false positive objects were deemed safety-critical by the baseline but not by the HJ safety zone. 
By using our mathematically complete, interaction-dynamics-aware safety zone, we observe significantly different results from the baseline model. 
We believe this difference stems from the strong theoretical assessment of safety-criticality that HJ reachability provides.
While additional closed-loop simulation experiments are needed to extensively quantify the benefits of our approach, and indeed, this is an avenue of future work (c.f. Section \ref{subsec:discussion:closed-loop-analysis}), the current results show great promise that our proposed safety zone leads to more informative safety-aware evaluation metrics.

\begin{table}[t]
\centering
\begin{tabular}{|C{2cm}||C{1.5cm}|C{1.6cm}|C{1.3cm}|}
\hline
\textbf{Method} & \textbf{Offline comp. time} & \textbf{Online comp. time} & \textbf{Memory usage}\\ \hline \hline
HJ safety zone & 228s & 4.5ms/object & 28.8M \\ \hline
Circular zone & --- & 0.07ms/object & ---\\ \hline
\end{tabular}
\caption{Comparison of computation cost}\label{tab:computation}
\end{table}

Lastly, a comparison of the computation costs for our HJ safety zones and the baseline method is shown in Table~\ref{tab:computation}. Although our safety zone requires some memory requirement to cache HJ value function, the online computation is still relatively lightweight, and therefore remains practical to use within a larger evaluation framework.

\section{Discussion \& Future Work} \label{sec:discussion}

In this section, we discuss interesting research directions that can extend our results.

\subsection{Generalization}

We have focused the majority of this work on a perception zone definition modeled after the false positive requirement introduced in Section~\ref{sec:problem-formulation}. A natural avenue of future work would be to extend our implementation to safety requirements targeting different dynamic scenarios and contender obstacle types. Examples of this might be to address cyclists and pedestrians, or to cover false negative detections. We believe that with the adjustable elements of the HJ reachability problem (see Section~\ref{subsec:adjustable_params}) our method is expressive enough to encompass a wide range of requirements.

\subsection{Closed-loop analysis} \label{subsec:discussion:closed-loop-analysis}

While we have argued that our approach is complete and theoretically superior to more na\"{i}ve approaches such as the circular baseline, demonstrating this empirically would require us to extend our current open-loop results with closed-loop experiments. This poses the interesting challenge that in order for our completeness property to hold, the planning and control modules used in a closed-loop setting would need to satisfy the modeling assumptions present in the safety zone definition. In the case of the false positive safety requirement used in this work, this entails that the downstream system would \emph{at worst} react no slower than 0.5s, and decelerate with at least 3.5ms$^{-2}$ until standstill.

\subsection{Temporal considerations} \label{subsec:discussion:temporal-considerations}

Throughout this work, we focused on single-frame analysis in our perception zone definitions. However, it is often expected that downstream planning and control modules are resilient to some amount of short, spurious perception failures. For a perception error to truly threaten a collision, safety-critical errors would likely need to last longer than some temporal threshold, or perhaps be distributed such that a resonance effect results in holistic system failure. In practice, our method can flag where safety-critical errors occur on a per-frame basis, but it may be desireable to augment the results with a model of the downstream system's temporal robustness to perception errors.

\subsection{Uncertainty in actor state}

It is often the case that an actor's state within a dynamic scenario is not known with full certainty, especially if it is being estimated by the system under test. This applies to the prototypical safety requirement that we focus on in this work, as it addresses false positive detections. 

Our HJ reachability-based method for defining zones lends itself well to accounting for uncertainty, and in fact allows for generalizing uncertainty into any dimensions of state space. If an actor's state is represented as a distribution instead of a point estimate, the corresponding safety-criticality information can be looked up by simply performing a weighted sum over the relevant region of the simulation grid.

\section{Conclusion}
\label{sec:conclusion}
We have presented an interaction-dynamics-aware perception zone to improve safety evaluation metrics for analyzing AV perception performance. The purpose of a perception zone is to help identify obstacles that pose a safety threat to the ego vehicle, i.e., are safety-critical, and use it to ground safety evaluation metrics.
By considering the dynamic states and closed-loop interactions between actors, our proposed perception zone is mathematically complete, and is more sound than other approaches that do not consider closed-loop interactions.
We leverage techniques from optimal control theory, namely HJ reachability, and existing computational tools to tractably compute such a interaction-dynamics-aware perception zone.
We performed experiments using the nuScene dataset and a baseline object detection algorithm and demonstrated that (i) our proposed interaction-dynamics-aware perception zone captures safety-critical false positive detections that are easily missed by a na\"{i}ve baseline, and (ii) by caching the perception zone solution offline, we can make our solution practically applicable to other perception evaluation tasks.

\bibliographystyle{IEEEtran}
\bibliography{bibfile}

\begin{thebibliography}{10}
\providecommand{\url}[1]{#1}
\csname url@samestyle\endcsname
\providecommand{\newblock}{\relax}
\providecommand{\bibinfo}[2]{#2}
\providecommand{\BIBentrySTDinterwordspacing}{\spaceskip=0pt\relax}
\providecommand{\BIBentryALTinterwordstretchfactor}{4}
\providecommand{\BIBentryALTinterwordspacing}{\spaceskip=\fontdimen2\font plus
\BIBentryALTinterwordstretchfactor\fontdimen3\font minus
  \fontdimen4\font\relax}
\providecommand{\BIBforeignlanguage}[2]{{%
\expandafter\ifx\csname l@#1\endcsname\relax
\typeout{** WARNING: IEEEtran.bst: No hyphenation pattern has been}%
\typeout{** loaded for the language `#1'. Using the pattern for}%
\typeout{** the default language instead.}%
\else
\language=\csname l@#1\endcsname
\fi
#2}}
\providecommand{\BIBdecl}{\relax}
\BIBdecl

\bibitem{iso2018road}
{International Organization for Standardization}, ``{{ISO 26262:2018}: Road
  Vehicles--Functional Safety},''
  \url{https://www.iso.org/standard/68391.html}, 2018.

\bibitem{albus20024d}
J.~S. Albus, ``{4D/RCS: A reference model architecture for intelligent unmanned
  ground vehicles},'' \emph{Unmanned Ground Vehicle Technology IV}, vol. 4715,
  pp. 303--310, 2002.

\bibitem{volk2020comprehensive}
G.~Volk, J.~Gamerdinger, A.~von Bernuth, and O.~Bringmann, ``{A Comprehensive
  Safety Metric to Evaluate Perception in Autonomous Systems},'' in
  \emph{{Proc.\ IEEE Int.\ Conf.\ on Intelligent Transportation Systems}},
  2020.

\bibitem{bansal2021risk}
A.~Bansal, J.~Singh, M.~Verucchi, M.~Caccamo, and L.~Sha, ``{Risk Ranked
  Recall: Collision Safety Metric for Object Detection Systems in Autonomous
  Vehicles},'' in \emph{{Mediterranean Conf.\ on Embedded Computing}}, 2021.

\bibitem{mitchell2005time}
I.~M. Mitchell, A.~M. Bayen, and C.~J. Tomlin, ``{A time-dependent
  {H}amilton-{J}acobi formulation of reachable sets for continuous dynamic
  games},'' \emph{{IEEE Transactions on Automatic Control}}, vol.~50, no.~7,
  pp. 947--957, 2005.

\bibitem{margellos2011hamilton}
K.~Margellos and J.~Lygeros, ``{{H}amilton--{J}acobi formulation for
  reach--avoid differential games},'' \emph{{IEEE Transactions on Automatic
  Control}}, vol.~56, no.~8, pp. 1849--1861, 2011.

\bibitem{bansal2017hamilton}
S.~Bansal, M.~Chen, S.~Herbert, and C.~J. Tomlin, ``{{H}amilton-{J}acobi
  reachability: A brief overview and recent advances},'' in \emph{{Proc.\ IEEE
  Conf.\ on Decision and Control}}, 2017.

\bibitem{nuscenes}
H.~Caesar, V.~Bankiti, A.~H. Lang, S.~Vora, V.~E. Liong, Q.~Xu, A.~Krishnan,
  Y.~Pan, G.~Baldan, and O.~Beijbom, ``nuscenes: A multimodal dataset for
  autonomous driving,'' in \emph{{IEEE Conf.\ on Computer Vision and Pattern
  Recognition}}, 2020, pp. 11\,621--11\,631.

\bibitem{dahl2019collision}
J.~Dahl, G.~Rodrigues~de Campos, C.~Olsson, and J.~Fredriksson, ``{Collision
  Avoidance: A Literature Review on Threat-Assessment Techniques},''
  \emph{{IEEE Transactions on Intelligent Vehicles}}, vol.~4, no.~1, pp.
  101--113, 2019.

\bibitem{aravantinos2020making}
V.~Aravantinos and P.~Schlicht, ``{Making the Relationship between Uncertainty
  Estimation and Safety Less Uncertain},'' in \emph{{ Conf.\ on Design,
  Automation and Test in Europe}}, 2020.

\bibitem{lyssenko2021evaluation}
M.~Lyssenko, C.~Gladisch, C.~Heinzemann, M.~Woehrle, and R.~Triebel, ``{From
  Evaluation to Verification: Towards Task-Oriented Relevance Metrics for
  Pedestrian Detection in Safety-Critical Domains},'' in \emph{{ IEEE/CVF Conf.
  on Computer Vision and Pattern Recognition Workshops}}, June 2021, pp.
  38--45.

\bibitem{hoss2021review}
M.~Hoss, M.~Scholtes, and L.~Eckstein. (2021) {A Review of Testing Object-Based
  Environment Perception for Safe Automated Driving}. {Available at
  }\url{https://arxiv.org/abs/2102.08460}.

\bibitem{philion2020learning}
J.~Philion, A.~Kar, and S.~Fidler, ``{Learning to Evaluate Perception Models
  Using Planner-Centric Metrics},'' in \emph{{IEEE Conf.\ on Computer Vision
  and Pattern Recognition}}, 2020.

\bibitem{Guo2020efficacy}
Y.~Guo, H.~Caesar, O.~Beijbom, J.~Philion, and S.~Fidler, ``{The efficacy of
  Neural Planning Metrics: A meta-analysis of PKL on nuScenes},'' in
  \emph{{IEEE/RSJ Int.\ Conf.\ on Intelligent Robots \& Systems: Workshop on
  Benchmarking Progress in Autonomous Driving}}, 2020.

\bibitem{berk2020assessing}
M.~Berk, O.~Schubert, H.-M. Kroll, B.~Buschardt, and D.~Straub, ``{Assessing
  the Safety of Environment Perception in Automated Driving Vehicles},''
  \emph{{SAE International Journal of Transportation Safety}}, vol.~8, no.~1,
  2020.

\bibitem{salay2019safety}
R.~Salay, M.~Angus, and K.~Czarnecki, ``A safety analysis method for perceptual
  components in automated driving,'' in \emph{{IEEE Int.\ Symp.\ on Software
  Reliability Engineering}}, 2019.

\bibitem{bajcsy2021towards}
A.~Bajcsy, K.~Leung, E.~Schmerling, and M.~Pavone. (2021) {Towards the
  Unification and Data-Driven Synthesis of Autonomous Vehicle Safety Concepts}.
  {Available at }\url{https://arxiv.org/abs/2107.14412}.

\bibitem{shalev2017formal}
S.~Shalev-Shwartz, S.~Shammah, and A.~Shashua. (2017) {On a formal model of
  safe and scalable self-driving cars}. {Available at
  }\url{https://arxiv.org/abs/1708.06374}.

\bibitem{bajcsy2019efficient}
A.~Bajcsy, S.~Bansal, E.~Bronstein, V.~Tolani, and C.~Tomlin, ``{An Efficient
  Reachability-Based Framework for Provably Safe Autonomous Navigation in
  Unknown Environments },'' in \emph{{Proc.\ IEEE Conf.\ on Decision and
  Control}}, 2019.

\bibitem{leung2020infusing}
K.~Leung, E.~Schmerling, M.~Zhang, M.~Chen, J.~Talbot, J.~C. Gerdes, and
  M.~Pavone, ``{On infusing reachability-based safety assurance within planning
  frameworks for human--robot vehicle interactions},'' \emph{{Int.\ Journal of
  Robotics Research}}, vol.~39, no. 10-11, pp. 1326--1345, 2020.

\bibitem{fisac2018general}
J.~F. Fisac, A.~K. Akametalu, M.~N. Zeilinger, S.~Kaynama, J.~Gillula, and
  C.~J. Tomlin, ``{A general safety framework for learning-based control in
  uncertain robotic systems},'' \emph{{IEEE Transactions on Automatic
  Control}}, vol.~64, no.~7, pp. 2737--2752, 2018.

\bibitem{HJJAX2021}
E.~Schmerling. {HJ Reachability in JAX}. {Available at
  }\url{https://github.com/StanfordASL/hj_reachability}.

\bibitem{vaskov2019not}
S.~Vaskov, H.~Larson, S.~Kousik, M.~Johnson-Roberson, and R.~Vasudevan,
  ``{Not-at-Fault Driving in Traffic: A Reachability-Based Approach},'' in
  \emph{{Proc.\ IEEE Int.\ Conf.\ on Intelligent Transportation Systems}},
  2019.

\bibitem{zhu2019class}
B.~Zhu, Z.~Jiang, X.~Zhou, Z.~Li, and G.~Yu. (2019) {Class-balanced grouping
  and sampling for point cloud 3D object detection}. {Available at
  }\url{https://arxiv.org/abs/1908.09492}.

\end{thebibliography}
\end{document}